%% file: 00_main.tex
\documentclass[10pt,twocolumn,letterpaper]{article}

\usepackage{iccv}
\usepackage{times}
\usepackage{epsfig}
\usepackage{graphicx}
\usepackage{amsmath}
\usepackage{amssymb}

\usepackage[pagebackref=true,breaklinks=true,letterpaper=true,colorlinks,bookmarks=false]{hyperref}

\iccvfinalcopy %

\def\iccvPaperID{9252} %

\ificcvfinal\fi
\input{00_definitions}

\begin{document}
\title{Stochastic Scene-Aware Motion Prediction}

\author{Mohamed Hassan$^1$ \; Duygu Ceylan$^2$ \; Ruben Villegas$^2$ \; Jun Saito$^2$ \; Jimei Yang$^2$ \; Yi Zhou$^2$ \; Michael Black$^1$\\
$^1$Max Planck Institute for Intelligent Systems, T{\"u}bingen, Germany \quad $^2$Adobe Research \\
{\tt\small \{mhassan, black\}@tue.mpg.de \quad \{ceylan, villegas, jsaito, jimyang, yizho\}@adobe.com}
}

\input{FIG_00_teaser}

\maketitle
\ificcvfinal\thispagestyle{empty}\fi
\begin{abstract}
A long-standing goal in computer vision is to capture, model, and realistically synthesize human behavior.
Specifically, by learning from data, our goal is to enable virtual humans to navigate within cluttered indoor scenes and naturally interact with objects.
Such embodied behavior has applications in virtual reality, computer games, and robotics, while synthesized behavior can be used as training data.
The problem is challenging because real human motion is diverse and adapts to the scene. For example, a person can sit or lie on a sofa in many places and with varying styles.
We must model this diversity to synthesize virtual humans that realistically perform human-scene interactions.
We present a novel data-driven, stochastic motion synthesis method that models different styles of performing a given action with a target object. Our Scene-Aware Motion Prediction method (SAMP) generalizes to target objects of various geometries while enabling the character to navigate in cluttered scenes. To train SAMP, we collected \mocap data covering various sitting, lying down, walking, and running styles. We demonstrate SAMP on complex indoor scenes and achieve superior performance than existing solutions. 
Code and data are available for research at \websiteURL.
\end{abstract}

\input{01_introduction}
\input{02_relatedwork}

\input{03_method}
\input{04_data}

\input{05_experiments}

\input{06_evaluation}

\input{07_conclusion}
{\small
\bibliographystyle{ieee_fullname}
\bibliography{00_bibliography}
}

\renewcommand{\thefigure}{S.\arabic{figure}}
\setcounter{figure}{0}
\renewcommand{\thetable}{S.\arabic{table}}
\setcounter{table}{0}

\clearpage
\begin{appendices}
\input{01_supmat}
\end{appendices}

\end{document}

%% file: 00_definitions.tex
\usepackage{times}
\usepackage{epsfig}
\usepackage{graphicx}
\usepackage{amsmath}
\usepackage{amssymb}

\usepackage{caption}
\usepackage{enumitem}
\usepackage{xspace}
\usepackage{subfigure}
\usepackage{hhline}
\usepackage[dvipsnames]{xcolor}
\usepackage{balance}
\usepackage{bm}
\usepackage[toc,page]{appendix}
\usepackage{multirow}
\usepackage[numbers,sort]{natbib}
\usepackage{mfirstuc}

\newcommand{\modelname}{\xspace{\mbox{SAMP}}\xspace}

\newcommand{\websiteURL}{\mbox{\url{https://samp.is.tue.mpg.de}}}

\newcommand{\smplX}{\xspace{\mbox{SMPL-X}}\xspace}

\newcommand{\supmat}{{\mbox{{Sup.~Mat.}}}\xspace}

\newcommand{\gt}{{ground truth}\xspace}

\newcommand{\mocap}{MoCap\xspace}

\renewcommand{\etc}{etc\xspace}
\renewcommand{\etal}{et al.\xspace}

\newcommand{\motionnet}{\xspace{MotionNet}\xspace}
\newcommand{\goalnet}{\xspace{GoalNet}\xspace}
\newcommand{\pathplanning}{\xspace{Path Planning Module}\xspace}
\newcommand{\gatingnet}{\xspace{Gating Network}\xspace}
\newcommand{\prednet}{\xspace{Prediction Network}\xspace}

\newcommand{\stateencoder}{\xspace{State Encoder}\xspace}
\newcommand{\interactionencoder}{\xspace{Interaction Encoder}\xspace}

\newcommand{\nsm}{\xspace{NSM}\xspace}

\newcommand{\shapenet}{\xspace{\mbox{ShapeNet}}\xspace}

\newcommand{\state}{\bm{X}}
\newcommand{\intersens}{\bm{I}}
\newcommand{\latentcode}{\bm{Z}}

\newcommand{\fd}{\xspace{Fr\`{e}chet distance}\xspace}

%% file: FIG_00_teaser.tex
\twocolumn[
{%
    \renewcommand\twocolumn[1][]{#1}%
    \maketitle
    \centering
\vspace{-0.26in}
    \begin{minipage}{1.00\textwidth}
    \centering				%
        \includegraphics[trim=000mm 000mm 000mm 000mm, clip=false, width=1.00 \textwidth]{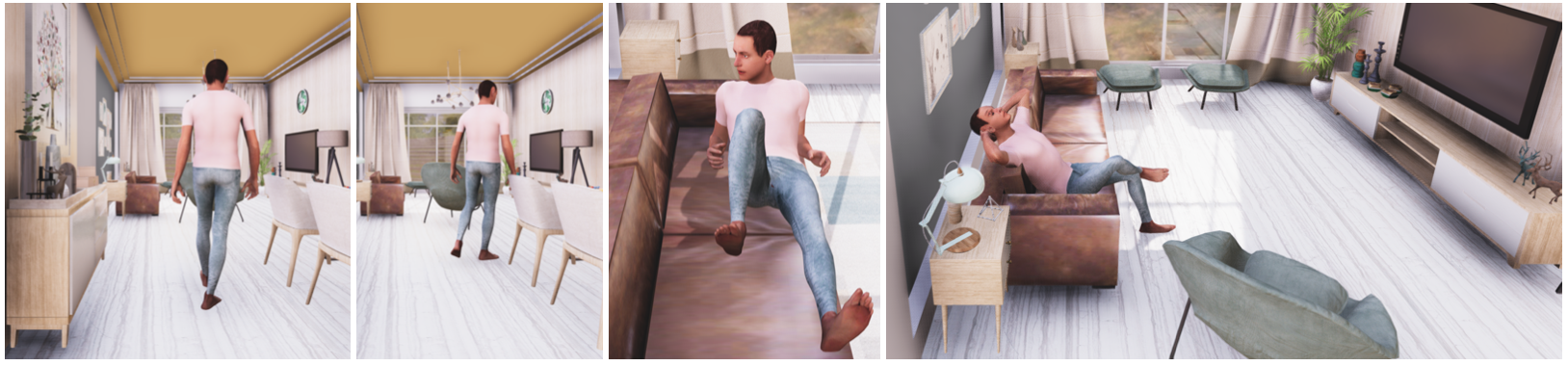}
    \end{minipage}
    \vspace{-0.1in}
    \captionof{figure}{
				\modelname synthesizes virtual humans navigating complex scenes with realistic and diverse human-scene interactions. 
    }\label{fig:teaser}
    \vspace*{+01.00em}
}
]%

%% file: 01_introduction.tex
\section{Introduction}
The computer vision community has made substantial progress on 3D scene understanding and on capturing 3D human motion, but less work has focused on synthesizing 3D people in 3D scenes. 
The advances in these two sub-fields, however, have provided tools for, and have created interest in,  embodied agents for virtual worlds (e.g.~\cite{habitat19iccv, xiazamirhe2018gibsonenv, shenigibson, xiang2020sapien}) and in placing humans into scenes (e.g.~\cite{cao2020long,hassan2020populating}).
Creating virtual humans that move and act like real people, however, is challenging and requires tackling many smaller but difficult problems such as perception of unseen environments, plausible human motion modeling, and embodied interaction with complex scenes. 
While  advances have been made in human locomotion modeling~\cite{Holden_2017,motion_vae_2020} thanks to the availability of large scale datasets~\cite{AMASS:ICCV:2019, TotalCapture_Trumble:BMVC:2017, cmuWEB, HEva_Sigal:IJCV:10b, MPI_HDM05},
realistically synthesizing virtual humans moving and interacting with 3D scenes, remains largely unsolved.

Imagine instructing a virtual human  to ``sit on a couch"  in a cluttered scene, as illustrated in Fig.~\ref{fig:teaser}. To achieve this goal, the character needs to perform a series of complex actions. First, it should navigate through the scene to reach the target object while avoiding collisions with other objects in the scene. Next, the character needs to choose a \emph{contact point} on the couch that will result in a plausible sitting action facing the right direction. %
Finally, if the character performs this action multiple times, there should be natural variations in the motion, mimicking real-world human-scene interactions;
e.g., sitting on different parts of the couch with different styles such as with crossed legs, arms in different poses, \etc.
Achieving these goals requires a system to jointly reason about the scene geometry, smoothly transition between cyclic (e.g., walking) and acyclic (e.g., sitting) motions, and to model the diversity of human-scene interactions.

To this end, we propose \modelname for \emph{Scene-Aware Motion Prediction}. 
\modelname is a stochastic model that takes a 3D scene as input, samples valid interaction goals, and generates goal-conditioned 
and scene-aware motion sequences of a character depicting realistic dynamic character-scene interactions. At the core of \modelname is a novel autoregressive conditional variational autoencoder (cVAE) called \motionnet. Given a target object and an action, \motionnet samples a random latent vector at each frame to condition the next pose both on the previous pose of the character as well as the random vector. This enables \motionnet to model a wide range of styles while performing the target action. 
Given the geometry of the target object, \modelname further uses another novel neural network called \goalnet to generate multiple plausible contact points and orientations on the target object (e.g., different positions and sitting orientations on the cushions of a sofa). This component enables \modelname to generalize across objects with diverse geometry. Finally, to ensure the character avoids obstacles while reaching the goal in a cluttered scene, we use an explicit path planning algorithm  (A* search) to pre-compute an obstacle-free path between the starting location of the character and the goal. This piecewise linear path consists of multiple way-points, which \modelname treats as intermediate goals to drive the character around the scene. \modelname runs in real-time at $30$ fps.
To the best of our knowledge, these individual components make \modelname the first system that addresses the problem of generating diverse dynamic motion sequences that depict realistic human-scene interactions in cluttered environments.

Training \modelname requires a dataset of rich and diverse character scene interactions. Existing large-scale \mocap datasets are largely dominated by locomotion and the few interaction examples lack diversity. Additionally, traditional \mocap focuses on the body and rarely captures the scene. Hence, we capture a new dataset covering various human-scene interactions with multiple objects. In each motion sequence, we track both the body motion and the object using a high resolution optical marker \mocap system. %
The dataset is available for research purposes.

Our contributions are: (1) A novel stochastic model for synthesizing varied goal-driven character-scene interactions in real-time. (2) A new method for modeling plausible action-dependent goal locations and orientations of the body given the target object geometry. (3) Incorporating explicit path planning  into a variational motion synthesis network enabling navigation in cluttered scenes. (4) A new \mocap dataset with diverse human-scene interactions.

%% file: 02_relatedwork.tex
\vspace{-0.05in}
\section{Related Work}
\vspace{-0.05in}
\textbf{Interaction Synthesis:}
Analyzing and synthesizing plausible human-scene interactions have received a lot of attention from the computer vision and graphics communities. Various algorithms have been proposed for predicting object functionalities~\cite{Grabner2011,zhu2015understanding}, affordance analysis~\cite{Gupta_2011,Wang_affordanceCVPR2017}, and synthesizing static human-scene interactions \cite{Grabner2011, Gupta_2011, hassan2020populating, Kim:2014:SHS, savva2016pigraphs, zhang2020place,zhang2020generating}. 

A less explored area involves generating dynamic human-scene interactions. While earlier work~\cite{Lee2002} focuses on synthesizing motions of a character in the same environment in which the motion was captured, follow up work~\cite{Lee_2006,Shum2008,2016-TOG-taskBasedLocomotion,Kapadia_2016} assembles motion sequences from a large database to synthesize interactions with new environments or characters. %
Such methods, however, require large databases and expensive nearest neighbor matching.

An important sub-category of human-scene interaction involves locomotion, where the character must respond to changes in terrain with appropriate foot placement. 
Phase-functioned neural networks~\cite{Holden_2017} have shown impressive results by using a guiding signal representing the state of the motion cycle (i.e.,~phase). %
Zhang \etal~\cite{mode_adaptive_zhang_2018} extend this idea to use a mixture of experts~\cite{Jacobs_MOE, eigen2013learning, Yuksel_MOE} as the motion prediction network. An additional gating network is used to predict the expert blending weights at run time. More closely related to our work is the Neural State Machine (NSM) \cite{nsm_2019}, which extends the ideas of phase labels and expert networks to model human-scene interactions such as sit, carry, and open. 
While NSM is a powerful method, it does not generate variations in such interactions, which is one of our key contributions.
Our experiments also demonstrate that NSM often fails to avoid intersections between the 3D character and objects in cluttered scenes (Sec.~\ref{sec:evaluation}). 
Furthermore, training NSM requires time-consuming manual, and often ambiguous, labeling of phases for non-periodic actions. Starke \etal~\cite{Starke_2020} propose a method to automatically extract local phase variables for each body part in the context of a two-player basketball game. Extending local phases to non-periodic actions is not trivial, however.
We find that using scheduled sampling \cite{Bengio_2015} provides an alternative to generate smooth transitions without phase labels.
More recently, Wang \etal~\cite{wang2020synthesizing} introduce a hierarchical framework for synthesizing human-scene interactions. They generate sub-goal positions in the scene, predict the pose at each of these sub-goals, and synthesize the motion between such poses. This method requires a post-optimization framework to ensure smoothness and robust foot contact and to discourage penetration with the scene.
Corona \etal~\cite{corona2020context} use a semantic graph to model human-object relationships followed by an RNN to predict human and object movements.

An alternative approach uses reinforcement learning (RL) to build a control policy that models interactions. Merel \etal~\cite{Merel_2020} and Eom \etal~\cite{Haegwang_Eom_2020} focus on ball catching from egocentric vision. %
Chao \etal~\cite{chao2019learning_to_sit} train sub-task controllers and a meta controller to execute the sub-tasks to complete a sitting task. However, in contrast to SAMP, their approach does not enable variations in the goal positions and directions. 
In addition, as with many RL-based approaches, generalizing the learned policies to new environments or actions is often challenging.

\textbf{Motion Synthesis:} Neural networks (feed-forward networks, LSTMs, or RNNs) have been extensively applied to the motion synthesis problem \cite{Fragkiadaki_2015, Habibie2017ARV, holden2016deep, martinez2017human, Taylor_2009, villegas2017learning, adeli2021tripod}. A typical approach predicts the future motion of a character based on previous frame(s). While showing impressive results when generating short sequences, many of these methods either converge to the mean pose or diverge when tested on long sequences. A common solution is to employ scheduled sampling \cite{Bengio_2015} to ensure stable predictions at test time to generate long locomotion and dancing sequences \cite{li2017auto, motion_vae_2020}. 

Several works have focused on modeling the stochastic nature of human motion, with a specific emphasis on trajectory prediction.
Given the past trajectory of a character, they model multiple plausible future trajectories \cite{cao2020long, MultiPath_2020, gupta2018social, MICB19, sadeghian2019sophie, Sidenbladh:ECCV:02, alahi2016social}.
Recently, Cao \etal~\cite{cao2020long} sample multiple future goals and then use them to generate different future skeletal motions. This is similar in spirit to our use of \goalnet. The difference is that our goal is to predict various trajectories that always lead to the same target object (instead of predicting any plausible future trajectory).

Modeling the stochasticity of the full human motion is a less explored area \cite{yuan2020dlow, Yuan2020Diverse, Wang_2021}. 
Motion VAE \cite{motion_vae_2020} predicts a distribution of the next poses instead of one pose using the latent space of a conditional variational auto-encoder. %
MoGlow is a controllable probabilistic generative model based on normalizing flows \cite{MoGlow}. 
Generating diverse dance motions from music has also been recently explored \cite{li2020learning, AISTPP_2021}.
Xu \etal~\cite{Xu_Hierarchical_2020} generate diverse motions by blending short sequences from a database.
To the best of our knowledge, no previous work has tackled the problem of generating diverse human-scene interactions.

%% file: 03_method.tex
\input{FIG_pipeline}
\section{Method}
Generating dynamic human scene interactions in cluttered environments requires solutions to several sub-problems. First and foremost, the synthesized motion of the character should be realistic and capture natural variations. Given a target object, it is important to sample plausible contact points and orientations for performing a specific action (e.g., where to sit on a chair and which direction to face). Finally, the motion needs to be synthesized such that it navigates to the goal location while avoiding penetrating objects in the scene. Our system consists of three main components that address each of these sub-problems: a \textit{\motionnet}, \textit{\goalnet}, and a \textit{\pathplanning}. At the core of our method is the \motionnet which predicts the pose of the character based on the previous pose as well as other factors such as the interaction object geometry and the target goal position and orientation. \goalnet predicts the goal position and orientation for the interaction on the desired object. The \pathplanning computes an obstacle-free path between the starting location of the character and the goal location. The full pipeline is illustrated in Fig.~\ref{fig:pipeline}.
\subsection{\motionnet} \label{sec:method_motion_prediction}
\input{03_method_01_motion_prediction}
\subsection{\goalnet} \label{sec:method_goalnet}
\input{03_method_02_goalnet}
\subsection{Path Planning} \label{sec:path_planning_avoidance}
\input{03_method_03_path_planning}
\subsection{Training Strategy} \label{sec:training_strategy}
\input{03_method_04_training_strategy}

%% file: FIG_pipeline.tex
\begin{figure*}
	\centering	
	\includegraphics[width=\linewidth]{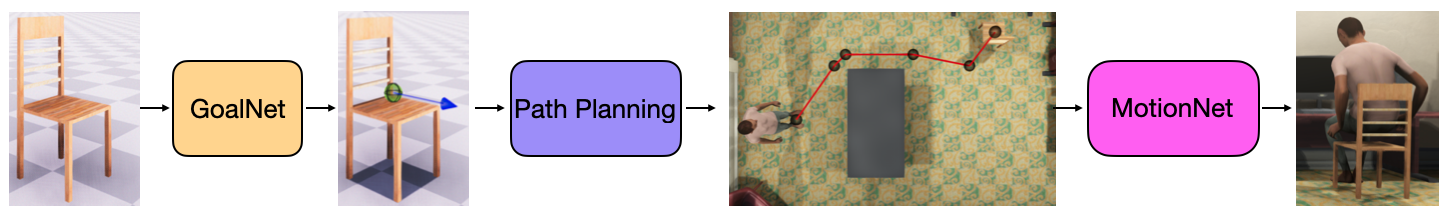}	
	\vspace{-0.3in}
	\caption{
	Our system consists of three main components. \goalnet predicts oriented goal locations (green sphere and blue arrow on the chair) given an interaction object. The \textit{\pathplanning} predicts an obstacle-free path from the starting position to the goal. \textit{\motionnet} sequentially predicts the next character state until the desired action is executed.
	}
	\label{fig:pipeline}
\end{figure*}

%% file: 03_method_01_motion_prediction.tex
\motionnet is an autoregressive conditional variational autoencoder (cVAE)~\cite{Kingma2014auto, cvae} that generates the pose of the character conditioned on its previous state (e.g., pose, trajectory, goal) as well as the geometry of the interaction object. \motionnet has two components: an encoder and a decoder. The encoder encodes the previous and current states of the character and the interaction object to a latent vector $\latentcode$.
The decoder takes this latent vector, the character's previous state, and the interaction object to predict the character's next state. The pipeline is shown in Fig.~\ref{fig:motionnet_arch}.  Note that, at test time, we only utilize the decoder of \motionnet and sample $\latentcode$ from a standard normal distribution.
\input{FIG_motionnet_arch}

\textbf{Encoder:}
The encoder consists of two sub-encoders: \textit{\stateencoder} and \textit{\interactionencoder}. 
The \stateencoder encodes the previous and current state of the character into a low-dimensional vector. Similarly, the \interactionencoder encodes the object geometry into a different low-dimensional vector. Next, the two vectors are concatenated and passed through two identical fully connected layers to predict the mean $\mu$ and standard deviation $\sigma$ of a Gaussian distribution representing a latent embedding space. We then sample a random latent code $\latentcode$, which is provided to the decoder when predicting the next state of the character. 

\textit{State Representation:} We use a representation similar to Starke \etal~\cite{nsm_2019} to encode the state of the character. Specifically, the state at frame $i$ is defined as $\state_i =$
\begin{equation}
    \left\{ \bm{j}^p_i,\bm{j}^r_i,\bm{j}^v_i, \Tilde{\bm{j}}^p_i, \bm{t}^p_i,\bm{t}^d_i,\Tilde{\bm{t}}^p_i,\Tilde{\bm{t}}^d_i,\bm{t}^a_i,\bm{g}^p_i, \bm{g}^d_i,\bm{g}^a_i, \bm{c}_i \right\},
    \label{eq:autoregression}
\end{equation}
where $\bm{j}^p_i \in \mathbb{R}^{3j},\bm{j}^r_i \in \mathbb{R}^{6j},\bm{j}^v_i \in \mathbb{R}^{3j}$ are the position, rotation, and velocity of each joint relative to the root. 
$j$ is the number of joints in the skeleton which is $22$ in our data. 
$\Tilde{\bm{j}}^p_i \in \mathbb{R}^{3j}$ are the joint positions relative to future root 1 second ahead.
$\bm{t}^p_i \in \mathbb{R}^{2t}$ and $\bm{t}^d_i \in \mathbb{R}^{2t}$ are the root positions and forward directions relative to the root of frame $i-1$.
$\Tilde{\bm{t}}^p_i \in \mathbb{R}^{2t}$ and $\Tilde{\bm{t}}^d_i \in \mathbb{R}^{2t}$ are the root positions and forward directions relative to the goal of frame $i-1$. We define these inputs for $t$ time steps sampled uniformly in a $2$ second window between $[-1,1]$ seconds. 
$\bm{t}^a_i \in \mathbb{R}^{n_a t}$ is a vector of continuous action labels on each of the $t$ samples.
In our experiments, $n_a$ is $5$, which is the total number of actions we model (i.e., idle, walk, run, sit, lie down). 
$\bm{g}^p_i \in \mathbb{R}^{3t}, \bm{g}^d_i \in \mathbb{R}^{3t}$ are the goal positions and directions, and $\bm{g}^a_i \in \mathbb{R}^{n_a t}$ is a one-hot action label describing the action to be performed at each of the $t$ samples. 
$\bm{c}_i \in \mathbb{R}^{5}$ are contact labels for pelvis, feet, and hands.

\textit{\stateencoder}: The \stateencoder takes the current $\state_i$ and previous state $\state_{i-1}$ and encodes them into a low-dimensional vector using three fully connected layers. 

\textit{\interactionencoder:} The \interactionencoder takes a voxel representation of the interaction object $\intersens$ and encodes it into a low-dimensional vector. We use a voxel grid of size $8 \times 8 \times 8$. Each voxel stores a $4-$dimensional vector. The first three components refer to the position of the voxel center relative to the root of the character. The fourth element stores the real-valued occupancy (between 0 and 1) of the voxel. The architecture consists of three fully connected layers.

\textbf{Decoder:}
The decoder takes the random latent code $\latentcode$, the interaction object representation $\intersens$, and the previous state $\state_{i-1}$, and predicts the next state $\hat{\state}_i$. Similar to recent work~\cite{motion_vae_2020, nsm_2019}, our decoder is built as a mixture-of-experts with two components: the \prednet and \gatingnet. 

The \prednet is responsible for predicting the next state $\hat{\state}_i$. The weights of the \prednet $\bm{\alpha}$ are computed by blending $K$ expert weights:
\begin{equation}
    \bm{\alpha} = \sum_{i=1}^{K} \omega_i \bm{\alpha}_i,
\end{equation}
where the blending weights $\omega_i$ are predicted by the \gatingnet. Each expert is a three-layer fully connected network.
The \gatingnet is also a three-layer fully connected network, which takes as input $\latentcode$ and $\state_{i-1}$.

\motionnet is trained end-to-end to minimize the loss $\mathcal{L}_\text{motion} = $
\begin{equation}
  || \hat{\state}_i - \state_i||_2^2 + \beta_\text{1} \mathit{KL}(Q(\latentcode|\state_i,\state_{i-1}, \intersens) || p(\latentcode)),
\end{equation}
where the first term minimizes the difference between the ground truth and predicted states of the character and $\mathit{KL}$ denotes the Kullback-Leibler divergence. 

%% file: FIG_motionnet_arch.tex
\begin{figure*}
	\centering	
	\includegraphics[width=0.8\linewidth]{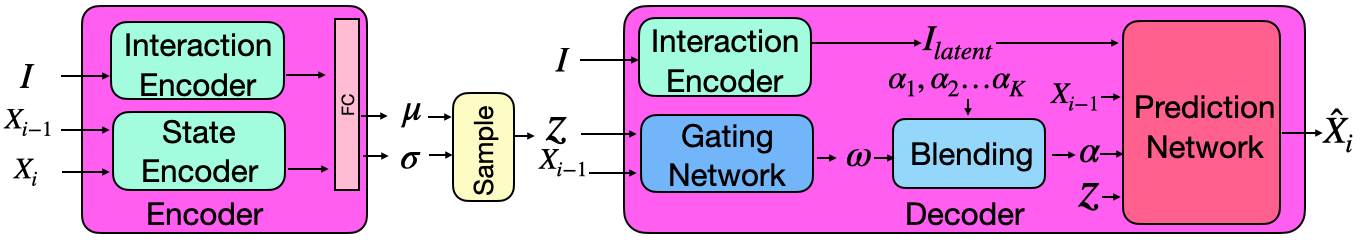}	
	\vspace{-0.1in}
	\caption{\motionnet consists of an encoder and a decoder. The encoder consists of two sub-encoders: \stateencoder and \interactionencoder. The decoder consists of a \prednet to predict the next character state and a gating network that predicts the blending weights of the \prednet. See Sec.~\ref{sec:method_motion_prediction}.
	}
	\label{fig:motionnet_arch}
	\vspace*{-01.00em}
\end{figure*}

%% file: 03_method_02_goalnet.tex
Given a target interaction object (which can be interactively defined by a user at test time or randomly sampled among the objects in the scene), the character is driven by the goal position $\bm{g}^p \in \mathbb{R}^{3}$ and direction $\bm{g}^d \in \mathbb{R}^{3}$ sampled on the object's surface. In order to perform realistic interactions; the character requires the ability to predict these goal positions and directions from the object geometry. 
For example, while a regular chair allows  variation in terms of sitting direction, the direction of sitting on an armchair is restricted (see Fig.~\ref{fig:gen_goals}).
We use \goalnet to model object-specific goal positions and directions. \goalnet is a conditional variational autoencoder (cVAE) that predicts plausible goal positions and directions given the voxel representation of the target interaction object $\intersens$ as shown in Fig.~\ref{fig:goalnet_arch}.  The encoder encodes the interaction object $\intersens$, goal position $\bm{g}^p$, and direction $\bm{g}^d$, into a latent code $\latentcode_{goal}$. The decoder reconstructs the goal position $\hat{\bm{g}}^p$, and direction $\hat{\bm{g}}^d$ from  $\latentcode_{goal}$ and $\intersens$.
We represent the object using a voxel representation similar to the one used in \motionnet(Sec.~\ref{sec:method_motion_prediction}). The only difference is that we compute the voxel position relative to the object center instead of the character root.
In the encoder, we use an \interactionencoder similar to the one used in \motionnet (see Sec.~\ref{sec:method_motion_prediction}) to encode the object representation $\intersens$ to a low dimension vector. This vector is then concatenated with $\bm{g}^p$ and $\bm{g}^d$ and encoded further to the latent vector $\latentcode_{goal}$. The decoder has the same architecture as the encoder as shown in Fig.~\ref{fig:goalnet_arch}. The network is trained to minimize the loss:
\begin{align}
    \mathcal{L}_\text{goal} =& || \hat{\bm{g}}^p - \bm{g}^p||_2^2 + || \hat{\bm{g}}^d - \bm{g}^d||_2^2 \nonumber \\ 
    & +\beta_2 \mathit{KL}(Q(\latentcode_{goal}|\bm{g}^p, \bm{g}^d, \intersens) || p(\latentcode_{goal})).
\end{align}
At test time, given a target object $\intersens$, we randomly sample $\latentcode_{goal} \sim \mathcal{N}(0,I)$ and use the decoder to generate various goal positions $\bm{g}^p$ and directions $\bm{g}^d$.
    
\input{FIG_goalnet_arch}

%% file: FIG_goalnet_arch.tex
\begin{figure}
	\centering	
	\includegraphics[width=\linewidth]{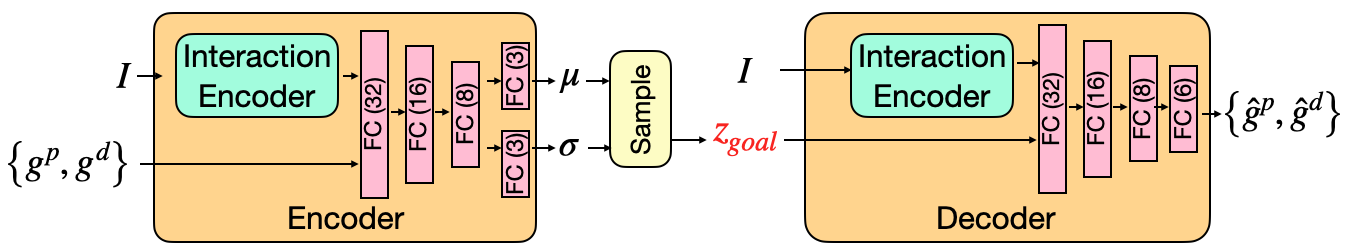}
	\vspace{-0.2in}
	\caption{\goalnet generates multiple valid goal positions $\hat{\bm{g}}^p$ and directions $\hat{\bm{g}}^d$ given an object representation $\intersens$.  FC(N) denotes a fully connected layer of size N.
	}
	\label{fig:goalnet_arch}
	\vspace*{-01.00em}
\end{figure}

%% file: 03_method_03_path_planning.tex
To ensure the character can navigate inside cluttered environments while avoiding obstacles, we employ an explicit A* path planning algorithm \cite{a_star}. 
Given the desired goal location, we use A* to compute an obstacle-free path from the starting position of the character to the goal. The path is defined as a series of waypoints $w_i = \left\{ w_0, w_1, w_2,...\right\}$ that define the locations where the path changes direction. We break the task of performing the final desired action into sub-tasks in which each sub-task requires the character to walk to the next waypoint. 
The final sub-task requires the character to perform the  desired action at the final waypoint.

%% file: 03_method_04_training_strategy.tex
Training \motionnet using standard supervised training produces poor quality predictions at run time (see \supmat). This is due to the accumulation of error at run time when the output of the network is fed back as input in the next step. To account for this, we train the network using \textit{scheduled sampling} \cite{Bengio_2015}, which has been shown to result in long stable motion predictions \cite{motion_vae_2020}. During training, the current network prediction is used as input in the next training step with a probability $1 - P$. $P$ is (see Sup.~Mat.): 
\begin{equation}
P = 
\begin{cases} 
1 & epoch \leq C_1, \\
1 - \frac{epoch - C_1}{C_2 - C_1} & C1 < epoch \leq C_2, \\
0 &  epoch >C2 .
\end{cases}
\end{equation}

%% file: 04_data.tex
\section{Data Preparation} \label{sec:data_preparation}
\subsection{Motion Data} \label{sec:motion_data_prep}
To model variations in human-scene interactions, we capture a new dataset using an optical \mocap system with $54$ Vicon cameras. We place seven different objects in the center of the \mocap area, namely two sofas, an armchair, a chair, a high bar chair, a low chair and a table. We record multiple clips of each interaction with different styles. %
In each sequence, the subject starts from an A-Pose in a random location in the \mocap space, walks towards the object, and performs the action for $20-40$ seconds. Finally, the subject gets up from the object and walks away. Our goal is to capture various styles of performing the same action, thus we ask the subject to change the style in each sequence. %
In addition to the subject, we also capture the object pose using attached markers. We also have the CAD model for each object. Finally, we capture running, walking, and idle sequences where the subject walks and runs in different directions with different speeds and stands in an idle state. 
Our dataset consists of ${\sim}100$ minutes of motion data recorded at $30$ fps from a single subject, resulting in ${\sim}185$K frames. 
We use MoSh++ \cite{AMASS:ICCV:2019} to fit the \smplX \cite{SMPL-X:2019} body model to the optical markers. 
More details about the data are available in the \supmat

\subsection{Motion Data Augmentation} \label{sec:motion_data_aug}
With only seven captured objects, \motionnet will fail to adapt to new unseen objects. Capturing \mocap with a wide range of objects  requires a significant amount of effort and time. We address this issue by augmenting our data using an efficient augmentation pipeline similar to \cite{Asqhar_2013, nsm_2019}. Since we capture both the body motion as well as the object pose, we compute the contact between the body and the object. We detect the contacts of five key joints of the character skeleton. Namely, pelvis, hands, and feet. We then augment our data by randomly switching or scaling the object at each frame. When switching, we replace the original object with a random object of a similar size selected from \shapenet~\cite{shapenet2015}. For each new object (scaled or switched), we project the contacts detected from the ground truth data to the new object. Finally, we use an IK solver to recompute the full pose such that the contacts are maintained. Please refer to the \supmat~for more details.

\subsection{Goal Data}
To train \goalnet, we label various goal positions $\bm{g}^p$ and directions $\bm{g}^d$ for different objects from \shapenet~\cite{shapenet2015}. These goals represent the position on the object surface where a character could sit and the forward direction of the character when sitting. We select $5$ categories from \shapenet namely, sofas, L-shaped sofas, chairs, armchairs, and tables. From each category, we select $15-20$ instances and we manually label $1-5$ goals for each instance. The number of goals labeled per instance depends on how many different goals an object can afford. For example, an L-shaped sofa offers more places to sit than a chair. In total, we use $80$ objects as our training data. We augment our data by randomly scaling the objects across the $xyz$ axes leading to ${\sim}13$K training samples.

%% file: 05_experiments.tex
\section{Experiments \& Evaluation}

\subsection{Qualitative Evaluation}
\label{sec:experiments}
In this section, we provide qualitative results and discuss the main points. We refer to the \supmat and the accompanying video for more results.

\textbf{Generating Diverse Motion:}\label{sec:experiments_diversity}
In contrast to previous deterministic methods~\cite{nsm_2019}, \modelname generates a wide range of diverse styles of an action while ensuring realism. Several different sitting and lying down styles generated by \modelname are shown in Fig.~\ref{fig:gen_styles}. The use of the \interactionencoder ~\ref{sec:method_motion_prediction} and the data augmentation (Sec.~\ref{sec:motion_data_aug}) further ensures \modelname can adapt  to different objects with varying geometry. Notice how the character naturally leans its head back on the sofa. The style of the action is also conditioned on the interacting object. The character lifts its legs when sitting on a high chair/table but extends its legs when sitting on a very low table. We observe that lying down is a harder task and several of baseline methods fail to execute this task (see Sec.~\ref{sec:evaluation}). While \modelname synthesizes reasonable sequences, our results are not always perfect. The generated motion might involve some penetration with the object. 
\input{FIG_gen_styles}

\textbf{Goal Generation:}\label{sec:experiments_goal}
When presented with a new object, the character needs to predict where and in which direction the action should be executed. 
In \cite{nsm_2019}, the goal is computed as the object center. 
However, this heuristic fails for objects with complex geometries. In Fig.~\ref{fig:ablation_goalnet} we show that using the object center results in invalid actions whereas \goalnet allows our method to reason about where the action should be executed. As shown in Fig.~\ref{fig:gen_goals}, by sampling different latent codes $\latentcode_{goal}$, \goalnet generates multiple goal positions and directions for various objects. Notice how \goalnet captures that, while a person can sit sideways on a regular chair, this is not valid for an armchair.
\input{FIG_ablation_goalnet}

\input{FIG_gen_goals}

Figure \ref{fig:gen_goal_motion} shows how the different goals generated by \goalnet  guide the motion of the character. Starting from the same position, direction, and initial pose, the virtual human follows two different paths to reach different goal positions when performing the ``sit on the couch" action. The final pose of the character is also different in the two cases due to the stochastic nature of \motionnet.
\input{FIG_gen_goals_motion}

\textbf{Path Planning:}
When navigating to a particular goal location in a cluttered scene, it is critical to avoid obstacles. Our \pathplanning achieves this goal by predicting the shortest obstacle-free path between the starting character position and the goal using a navigation mesh computed based on the 3D scene. %
The navigation mesh defines the walk-able areas in the scene and is computed once offline. In Fig.~\ref{fig:ablation_pathplanning}, we show an example path computed by the \pathplanning. Without this module, the character often walks through objects in the scene. We observe a similar behaviour in the previous work of \nsm~\cite{nsm_2019}, even though \nsm uses a volumetric representation of the environment to help the character navigate.
\input{FIG_ablation_path_planning}

%% file: FIG_gen_styles.tex
\begin{figure*}[ht]
	\centering	
	\includegraphics[width=\linewidth]{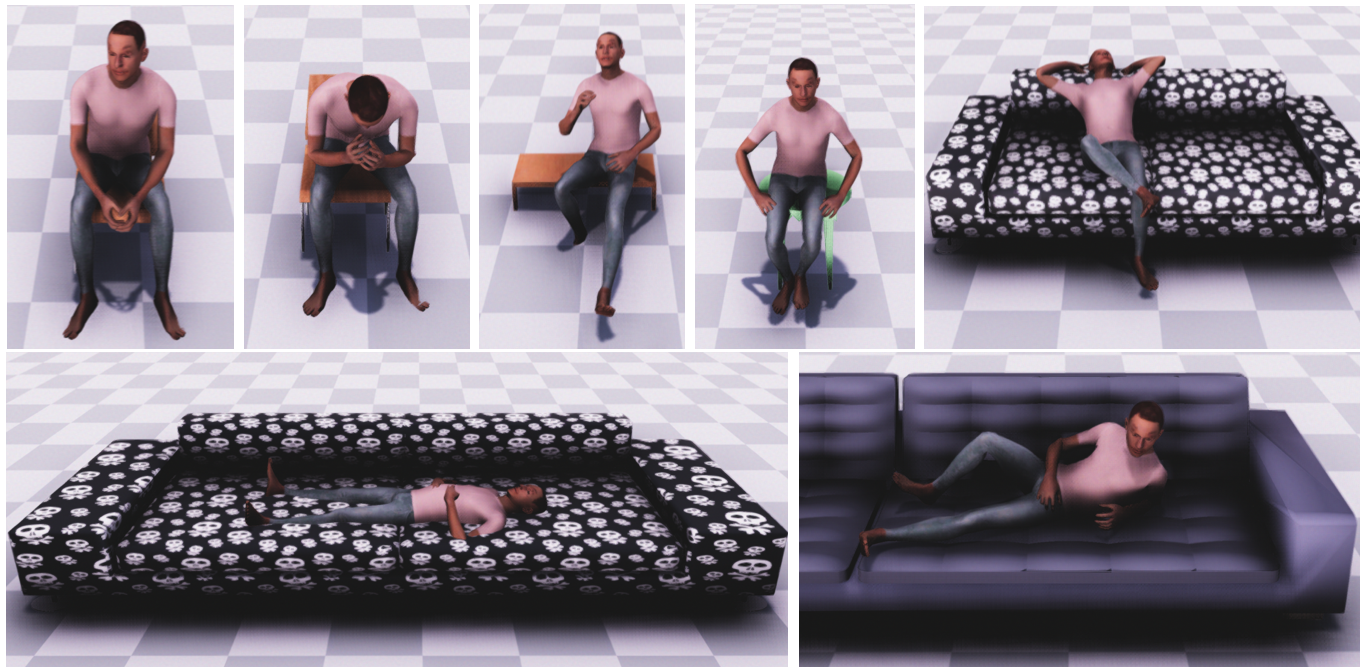}	
	\vspace{-0.2in}
	\caption{\modelname generates plausible and diverse action styles and adapts to different object geometries. 
	}
	\label{fig:gen_styles}
\end{figure*}

%% file: FIG_ablation_goalnet.tex
\begin{figure}
	\centering	
	\includegraphics[width=\linewidth]{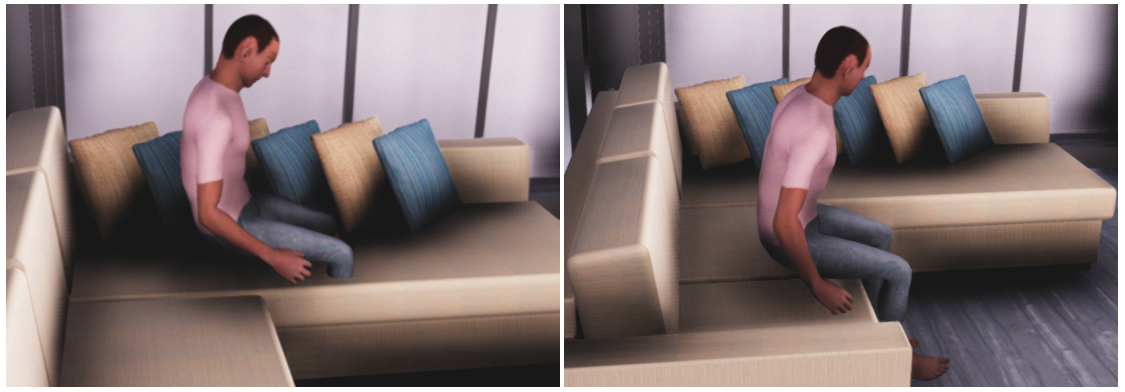}	
	\vspace{-0.2in}
	\caption{
	Without \goalnet (left), \modelname fails to sit on a valid place. \modelname with \goalnet is shown on the right.
	}
	\label{fig:ablation_goalnet}
\end{figure}

%% file: FIG_gen_goals.tex
\begin{figure}
	\centering	
	\includegraphics[width=\linewidth]{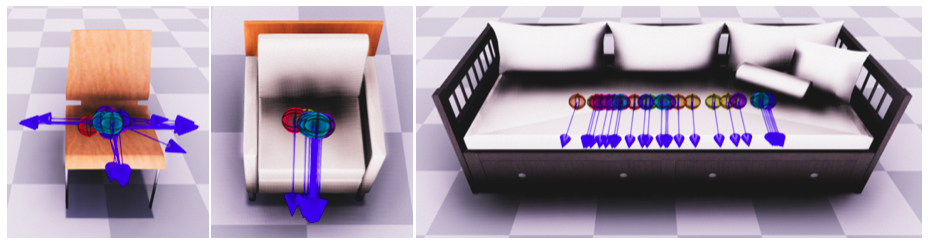}
	\vspace{-0.2in}
	\caption{
	\goalnet generates diverse valid goals on different objects. Spheres indicate goal positions, and blue arrows indicate goal directions.
	}
	\label{fig:gen_goals}
	\vspace*{-01.00em}
\end{figure}

%% file: FIG_gen_goals_motion.tex
\begin{figure*}[t]
	\centering	
	\includegraphics[width=\linewidth]{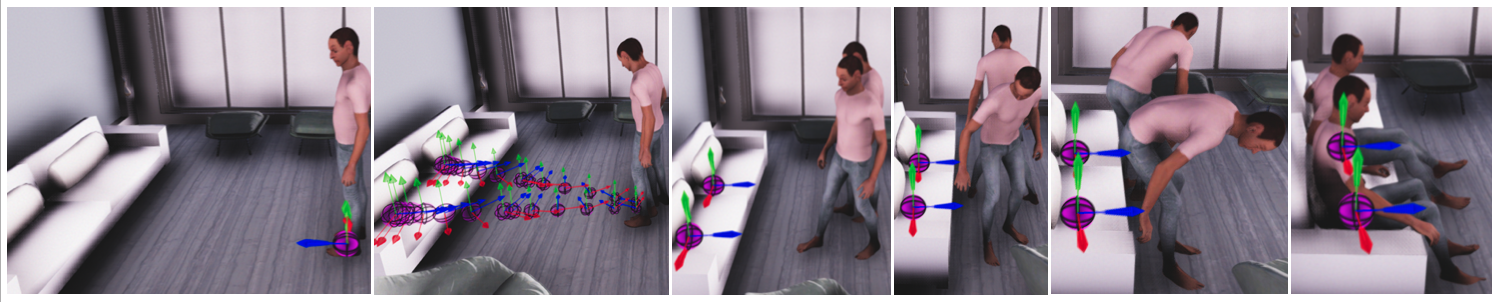}	
	\vspace{-0.28in}
	\caption{
	Goals generated by \goalnet (mesh spheres) are used by \motionnet to guide the motion of virtual characters. 
	}
	\label{fig:gen_goal_motion}
\end{figure*}

%% file: FIG_ablation_path_planning.tex
\begin{figure}
	\centering	
	\includegraphics[width=\linewidth]{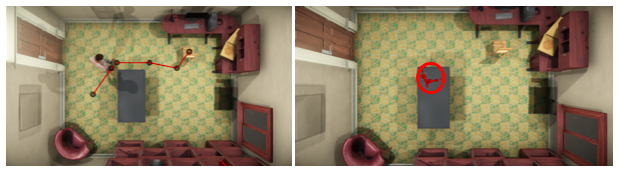}	
	\vspace{-0.27in}
	\caption{
	Our \pathplanning helps \modelname to successfully navigate cluttered scenes (left). NSM~\cite{nsm_2019} fails in such scenes (right).
	}
	\label{fig:ablation_pathplanning}
\end{figure}

%% file: 06_evaluation.tex
\subsection{Quantitative Evaluation}
\label{sec:evaluation}
\textbf{Deterministic vs. Stochastic:} To quantify the diversity of the generated motion, we put the character in a fixed starting position and direction and we run our method ten times with the same goal. For example, we instruct the character to sit/lie down on the same object multiple times starting from the same initial state/position/direction. For walking and running, we instruct the character to run in each of the four directions for $15$ seconds. We record the character motion for each run and then compute the Average Pairwise Distance (APD) \cite{yuan2020dlow, zhang2020we} as shown in Table.~\ref{tab:diverstity}. The APD is defined as:
\begin{equation}
    APD = \frac{1}{N(N-1)} \sum_{i=0}^{N} \sum_{\substack{j = 0\\ j\neq i}}^{N} || {\state}'_i - {\state}'_j||_2^2.
\end{equation}
${\state}'_i$ represents the character's local pose features at frame $i$. ${\state}'_i = \left\{ \bm{j}^p_i,\bm{j}^r_i,\bm{j}^v_i\right\}$. $N$ is the total number of frames for all sequences. For comparison we also report the APD for the \gt (GT) data in Table.~\ref{tab:diverstity}.
\input{TAB_diversity}

\textbf{\goalnet:} Given $150$ unseen goals sampled on test objects, we measure the average position and orientation reconstruction error of \goalnet to be $\bm{6.04}$ cm and $\bm{2.29}$ deg (we note that the objects have real-life measurements). To measure the diversity of the generated goals, we compute the Average Pairwise Distance (APD) among the generated goal positions $g^p$ and directions $g^d$: 
\begin{align}
    \text{APD-Pos} &= \frac{1}{L  N  (N-1)} \sum_{k=0}^{L} \sum_{i=0}^{N} \sum_{\substack{j = 0\\ j\neq i}}^{N} | g^p_i - g^p_j| \\
    \text{APD-Rot} &= \frac{1}{L N  (N-1)} \sum_{k=0}^{L} \sum_{i=0}^{N} \sum_{\substack{j = 0\\ j\neq i}}^{N} \arccos(g^d_i . g^d_j).
\end{align}
$L=150$ is the number of objects and $N=10$ is the number of goals generated for each object. We find APD-Pos and APD-Rot for our generated goals to be $\bm{16.42}$ cm and $\bm{41.27}$ deg compared to $\bm{16.18}$ cm and $\bm{90.23}$ deg for the \gt(GT) data.

\textbf{\pathplanning:} To quantitatively evaluate the effectiveness of our \pathplanning, we test our method in a cluttered scene. We put the character in a random initial position and orientation and select a random goal. We repeat this  $10$ times. We find the percentage of frames where a penetration happens is $\bm{3.8\%}$,  $\bm{11.2\%}$, and $\bm{8.11\%}$ for \modelname with \pathplanning, without \pathplanning, and \nsm ~\cite{nsm_2019}, respectively. While \nsm uses a volumetric sensor to detect collisions with the environment, it is not as effective as explicit path planning.

\textbf{Comparison to Previous Models:} We compare our model to baselines by measuring three metrics: average execution time, average precision, and \fd (FD) between the distribution of the generated motion and \gt. Execution time is the time required to transition to the target action label from an idle state. Precision is the positional (PE) and rotational (RE) error at the goal. We measure FD on a subset of the state features which we call $\Tilde{\state}$:
\begin{equation}
    \Tilde{\state} = \left\{ \bm{j}^p,\bm{j}^r,\bm{j}^v, \Tilde{\bm{t}}^p,\Tilde{\bm{t}}^d \right\}.
\end{equation}
As our baselines, we choose a feedforward network (MLP) as the motion prediction network, Mixture of Experts (MoE)~\cite{mode_adaptive_zhang_2018}, and NSM~\cite{nsm_2019} (see \supmat for details).

\textit{\modelname vs.~MLP vs.~MoE:}
We re-trained the MLP and MoE using the same training strategy and data we used for \modelname. Both MLP and MoE take a longer time to execute the task and often fail to execute the ``lie down" action (denoted $\infty$) as evidenced by the execution time in Table.~\ref{tab:execution_time} and precision in Table.~\ref{tab:precision}. These architectures sometimes generate implausible poses as shown in \supmat, which is reflected by the lower FD in Table.~\ref{tab:fd_1}

\input{TAB_execution_time}

\input{TAB_precision}

\input{TAB_FD_1}

\textit{\modelname vs.~NSM:}
For NSM, we used the publicly available pre-trained model since retraining NSM on our data is infeasible due to the missing phase labels. We trained \modelname on the same data on which NSM was trained. In Table~\ref{tab:NSM_comparison} we observe that our model is on par with \nsm in terms of achieving goals without the need for phase labels, which are cumbersome and often ambiguous to annotate. In addition, our main focus is to model diverse motions via a stochastic model while \nsm is deterministic. 
Our \pathplanning module helps \modelname to safely navigate complex scenes where \nsm fails as shown by the penetration amounts.

For all evaluations, all test objects are randomly selected from \shapenet and none is part of our training set.
\input{TAB_NSM_precision}

\textbf{Limitations and Future Work:} We observe that sometimes slight penetrations between the character and the interacting object can occur. A potential solution is to incorporate a post-processing step to optimize the pose of the character to avoid such intersections. In order to generalize \modelname to interacting objects that have significantly different geometry than those seen in training, in future work, we would like to explore methods to encode local object geometries.

%% file: TAB_diversity.tex
\begin{table}[]
\centering
\begin{tabular}{|c|c|c|c|c|}
\hline
 & Walk & Run & Sit & Liedown \\ \hline
GT &     $5.95$ & $7.74$  &   $5.18$   & $7.52$     \\ \hline
\modelname & $5.63$ & $5.75$ & $5.05$& $6.69$ \\ \hline
\end{tabular}
\caption{Diversity metric. Higher values indicate more diversity.}
\label{tab:diverstity}
\vspace*{-01.00em}
\end{table}

%% file: TAB_execution_time.tex
\begin{table}[]
\centering

\begin{tabular}{|c|c|c|c|c|}
\hline
    & MLP & MoE  & \modelname & GT \\ \hline
Sit       &     $13.06$         &   $12.99$  & $\boldsymbol{12.53}$ &  $11.7$ \\ \hline
Liedown       &     $\infty$        & $\infty$ &    $\boldsymbol{17.06}$ &    $15.49$  \\ \hline
\end{tabular}

\caption{Average execution Time in seconds. $\infty$ means the method failed to reach the goal within 3 minutes.}
\label{tab:execution_time}
\end{table}

%% file: TAB_precision.tex
\begin{table}[]
\centering
\begin{tabular}{|c|c|c|c|c|}
\hline
\multirow{2}{*}{Method}   & \multicolumn{2}{c|}{Sit} & \multicolumn{2}{c|}{Liedown} \\ \cline{2-5} 
                          & PE(cm)     & RE(deg)     & PE(cm)        & RE(deg)      \\ \hline
MLP                       & $9.27$      & $3.99$        & $\infty$      & $\infty$     \\ \hline
MoE                       & $7.99$       & $5.73$        & $\infty$      & $\infty$     \\ \hline
\modelname                & $\boldsymbol{6.09}$       & $\boldsymbol{3.55}$        & $\boldsymbol{5.76}$          & $\boldsymbol{6.45}$         \\ \hline
\end{tabular}
\caption{Average precision in terms of positional and rotational errors (PE and RE). $\infty$ means the method failed to reach the goal within 3 minutes.}
\label{tab:precision}
\end{table}

%% file: TAB_FD_1.tex
\begin{table}[]
\centering
\resizebox{\linewidth}{!}
{
\begin{tabular}{|c|c|c|c|c|c|}
\hline
            & Idle      & Walk      & Run     & Sit         & Liedown \\ \hline
MLP         & $102.85$    & $121.18$    & $150.56$  & $105.87$      & $36.85$     \\ \hline
MoE         & $102.91$    & $114.17$    & $151.14$  & $105.10$      & $35.79$     \\ \hline
\modelname  & $\boldsymbol{102.72}$    & $\boldsymbol{111.09}$    & $\boldsymbol{141.11}$  & $\boldsymbol{104.68}$      & $\boldsymbol{17.30}$ \\ \hline
\end{tabular}
}
\caption{\fd.}
\label{tab:fd_1}
\vspace*{-01.00em}
\end{table}

%% file: TAB_NSM_precision.tex
\begin{table}[]
\small
\centering
\begin{tabular}{|c|c|c|c|c|}
\hline
\multirow{2}{*}{Metric} & \multicolumn{2}{c|}{Sit} & \multicolumn{2}{c|}{Carry} \\ \cline{2-5} 
                        & SAMP        & NSM        & SAMP         & NSM         \\ \hline
Precision PE (cm)  $\downarrow$      &     $\boldsymbol{15.97}$        &   $16.95$  &         $\boldsymbol{4.58}$     &     $4.72$        \\ \hline
Precision RE (deg) $\downarrow$      &     $5.38$        &    $\boldsymbol{2.32}$        &         $1.78$     &        $\boldsymbol{1.65}$     \\ \hline
Execution Time (sec) $\downarrow$         &   $12.93$          &   $\boldsymbol{10.26}$   &         $13.29$     &   $\boldsymbol{12.82}$          \\ \hline
FD  $\downarrow$             &      $6.20$       &    $\boldsymbol{4.21}$        &       $10.17$       &    $\boldsymbol{7.31}$         \\ \hline
Diversity  $\uparrow$             &      $\boldsymbol{0.44}$       &    $0.0$        &       $\boldsymbol{0.26}$       &    $0.0$         \\ \hline
Penetration ($\%$)   $\downarrow$           &     $\boldsymbol{3.8}$        &      $8.11$      &     $\boldsymbol{3.62}$        &    $8.45$        \\ \hline
\end{tabular}
\caption{SAMP vs.~NSM.}
\label{tab:NSM_comparison}
\vspace{-2.0 em}
\end{table}

%% file: 07_conclusion.tex
\section{Conclusion}
Here we have described \modelname, which makes several important steps toward creating lifelike avatars that move and act like real people in previously unseen and complex environments. Critically, we introduce three elements that must be part of a solution.
First, characters must be able to navigate the world and avoid obstacles.  For this, we use an existing path planning method.
Second, characters can interact with objects in different ways.  To address this, we train \goalnet to take an object and stochastically produce an interaction location and direction.  
Third, the character should produce motions achieving the goal that vary naturally. 
To that end, we train a novel \motionnet that incrementally generates body poses based on the past motion and the goal.
We train \modelname using a novel dataset of motion capture data involving human-object interaction. %

{
\noindent
\small
\textbf{Acknowledgement}
This work was initiated while MH was an intern at Adobe. We are grateful to Sebastian Starke for inspiring work, helpful discussion, and making his code open-source. We thank Joachim Tesch for feedback on Unity and rendering, Nima Ghorbani for MoSH++, and Meshcapade for the character texture. For helping with the data collection, we are grateful to Tsvetelina Alexiadis, Galina Henz, Markus H\"{o}schle and Tobias Bauch.

\noindent
\textbf{Disclosure:}
MJB has received research funds from Adobe, Intel, Nvidia, Facebook, and Amazon. While MJB is a part-time employee of Amazon, his research was performed solely at, and funded solely by, Max Planck. MJB has financial interests in Amazon, Datagen Technologies, and Meshcapade GmbH.
}

%% file: 01_supmat.tex
\section{Data Preparation}
\subsection{Motion Data}
Fig.~\ref{fig:mocap} shows examples of different sitting and lying down styles from our \mocap. A breakdown of the dataset in terms of different actions is shown in Table~\ref{tab:data_breakdown}. The objects used during the \mocap are shown in Fig.~\ref{fig:Objects}.
\input{FIG_mocap}
\input{FIG_objects}
\input{TAB_data_breakdown}

\subsection{Goal Data}
We select $5$ categories from \shapenet namely, sofas, L-shaped sofas, chairs, armchairs, and tables. From each category, we select $15-20$ instances and we manually label $1-5$ goals for each instance. Table.~\ref{tab:goalnet_data_breakdown} shows the number of instances for each category.
We manually label $1-5$ goals for each instance. The number of goals labelled per instance depends on how many different goals an object can afford. For example, we label $5$ different goals for the L-shaped sofa compared to $3$ for the chair as shown in Fig.~\ref{fig:goal_labelling}.
\input{FIG_goal_labelling}
\input{TAB_goalnet_data}

 \section{Training Details}
 \subsection{\motionnet}
 The character state $\state$ is of size $647$. The \stateencoder, \interactionencoder, \gatingnet, and \prednet are all three-layer fully connected networks with rectified linear function ELU. The dimensions of each network are in Table~\ref{tab:arch}. The encoder latent code $\latentcode$ is of size $64$ and we set the number of experts $K$ to $12$.
 We use a learning rate of $5e-5$ and train our network for $100$ epochs. We use the Adam optimizer with linear weight decay. The weight of the Kullback-Leibler divergence $\beta_\text{1}$ is $0.1$.
 \input{TAB_arch}

\subsection{\goalnet}
The \interactionencoder of \goalnet is a three-layer fully connected network of shape $\left\{ 512, 512, 64 \right\}$. The latent vector $\latentcode_{goal}$ is of size $3$. The weight of the Kullback-Leibler divergence $\beta_2$ is $0.5$. We use the Adam optimizer with a learning rate of $1e-3$ and train \goalnet for $100$ epochs.

 \subsection{Schedule Sampling}
 For the schedule sampling training strategy, we set $C_1=30$ and $C_2=60$. We define a roll-out window of size $L$ where we set $L=60$ in our experiments. For each roll-out, we feed the ground truth first frame as input to the network and then sequentially predict the subsequent frames while using the scheduled sampling strategy. We divide our training data to equal-length clips of size $L$.
 
 \section{Baselines}
As our baselines, we choose a feedforward network (MLP) and a Mixture of Experts (MoE). The architecture of the MLP is shown in Fig.~\ref{fig:mlp}. We use the same \interactionencoder used for our \motionnet followed by four fully connected layers of size $512$. The architecture of the MoE is shown in Fig.~\ref{fig:moe}. The \interactionencoder, \gatingnet, and \prednet are all the same as the one used in \motionnet.
\input{FIG_MLP}
\input{FIG_MoE}

\section{Schedule Sampling} We found that using Schedule Sampling is essential to enable the character to successfully reach the goal and execute the action. Without it, we found the model to often diverge, get stuck, or take very long time to reach the goal as we show in Fig.~\ref{fig:sched_sampling}.
\input{FIG_shced_sampling}

\section{Path Planning Formulation} 
In order to use the \pathplanning, we first compute the surface area where the character could stand or move. We call this the \textit{navigation mesh}. This is computed from the character cylinder collider and the scene geometry. The \textit{navigation mesh} is stored as convex polygons. To find a path between given start and end points, we first map these points to the closest polygons and then use A* to find the shortest path between the polygons \footnote{\url{https://docs.unity3d.com/Manual/nav-InnerWorkings.html}}. 

\section{Data Augmentation Details}
When the object is transformed, the contacts follow the same transformation. When the object is replaced by a new one, we project the original contact by finding the closest points on the surface of the new object. The new motion curve is computed by interpolation and the whole full body pose is computed using CCD IK solver. This does not guarantee smoothness but we found it to be stable in practice. More details are in \cite{nsm_2019}.

\section{\interactionencoder Ablation:}
To quantify the importance of the \interactionencoder, we trained \modelname without the \interactionencoder. We found that the precision of reaching the goal deteriorates to $14.82$ cm and $3.65$ deg compared to $6.09$ cm and $3.55$ deg when the \interactionencoder was used.

\section{Comparison to~Cao \etal:}
While relevant, the formulation of Cao~\cite{cao2020long}  \etal~is significantly different than our method making a direct comparison difficult. Given a target interaction object and action (e.g.~``sit on the couch"), \modelname samples a goal location and orientation on the object, computes an obstacle-free path towards the object, and synthesizes diverse motion sequences that are of arbitrary length until the goal is executed. We assume that the character starts the action from an idle position without any knowledge of the past. In contrast, Cao \etal~sample a goal {\em location} in the image space given a one-second-long {\em history} of motion. Based on this trajectory, a deterministic motion sequence of fixed length (two-seconds) is synthesized. The action executed in this trajectory is not controllable. 

\section{Failure Cases}
We observe that \modelname might not adapt well to objects with significantly different geometry than those seen in training as shown in Fig.~\ref{fig:new_goemetry}. Future work might explore different methods of encoding the object geometry. 
\input{FIG_new_geometry}

%% file: FIG_mocap.tex
\begin{figure}[h!]
	\centering	
	\includegraphics[width=\linewidth]{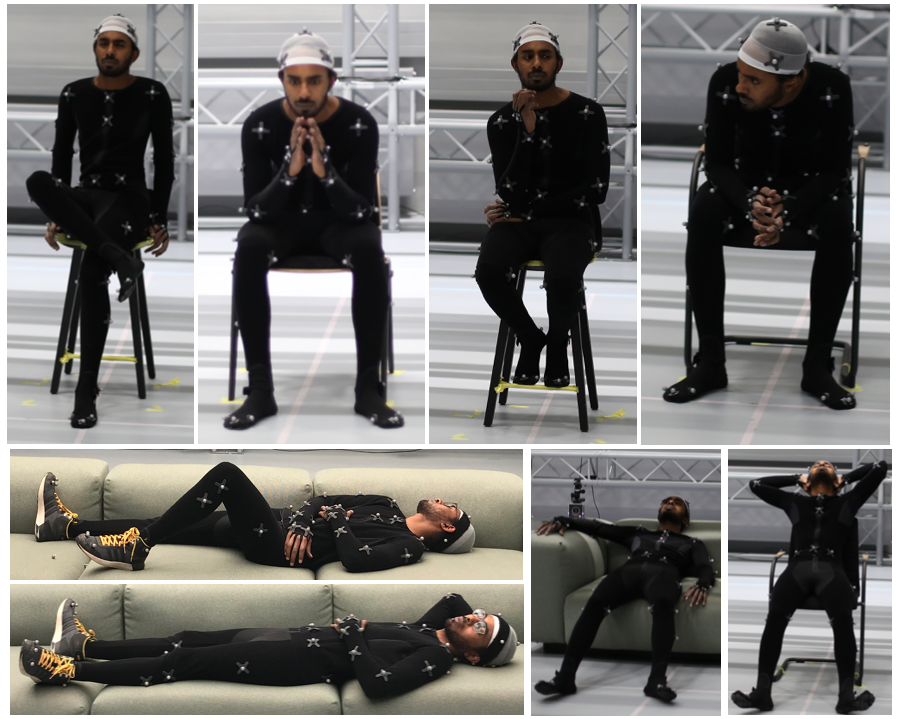}	
	\caption{
	Examples of action styles in our motion capture data.
	}
	\label{fig:mocap}
\end{figure}

%% file: FIG_objects.tex
\begin{figure}[h!]
	\centering	
	\includegraphics[width=\linewidth]{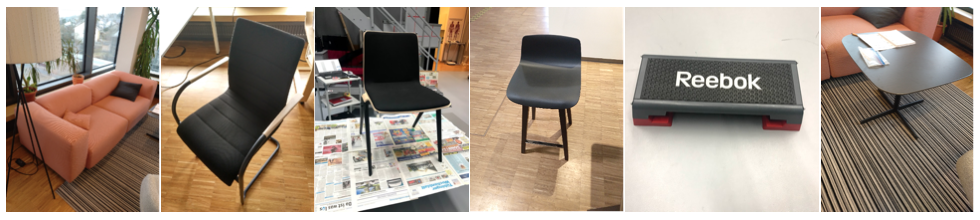}	
	\caption{
	Objects used during motion capture.
	}
	\label{fig:Objects}
\end{figure}

%% file: TAB_data_breakdown.tex
\begin{table}[h!]
\centering
\begin{tabular}{c|c|c}
\hline
Labels   & Minutes & Percentage \% \\ \hline
Idle     &  $18.3$    &   $17.7$ \\
Walk     &  $42.3$    &   $41.0$ \\
Run      &  $5.1$     &    $4.9$ \\
Sit      &    $27.3$  &    $26.4$\\
Lie down &       $10.1$  &    $9.7$\\
\textbf{Total}    & $103.3$ & $\-$\\ \hline
\end{tabular}
\caption{Motion capture data breakdown with respect to actions.}
\label{tab:data_breakdown}
\end{table}

%% file: FIG_goal_labelling.tex
\begin{figure}
	\centering	
	\includegraphics[width=\linewidth]{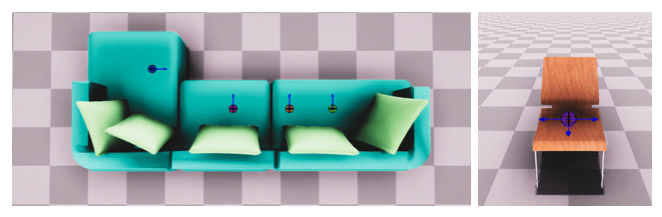}	
	\caption{
	Goal Labelling.
	}
	\label{fig:goal_labelling}
\end{figure}

%% file: TAB_goalnet_data.tex
\begin{table}[]
\centering
\begin{tabular}{c|c}
\hline
Category   & Number of Objects\\ \hline
Armchairs     &  $15$     \\
Chairs     &  $16$     \\
Sofa      &  $20$      \\
L-Sofa      &    $18$  \\
Tables &       $18$  \\
\textbf{Total}    & $87$ \\ \hline
\end{tabular}
\caption{\goalnet data breakdown with respect to object categories.}
\label{tab:goalnet_data_breakdown}
\vspace*{-01.00em}
\end{table}

%% file: TAB_arch.tex
\begin{table}[]
\centering
\begin{tabular}{c|c}
\hline
Network   & Architecture\\ \hline
\stateencoder    &  $\left\{ 512, 256, 256 \right\}$     \\
\interactionencoder      &  $\left\{ 256, 256, 256 \right\}$     \\
\gatingnet & $\left\{ 512, 256, 12 \right\}$     \\
\prednet & $\left\{ 512, 512, 647 \right\}$     \\
\end{tabular}
\caption{Architecture details. All networks are all three-layer fully connected networks with ELU.}
\label{tab:arch}
\end{table}

%% file: FIG_MLP.tex
\begin{figure}
	\centering	
	\includegraphics[width=\linewidth]{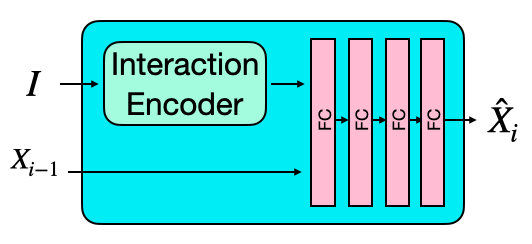}	
	\caption{
	MLP Architecture.
	}
	\label{fig:mlp}
\end{figure}

%% file: FIG_MoE.tex
\begin{figure}
	\centering	
	\includegraphics[width=\linewidth]{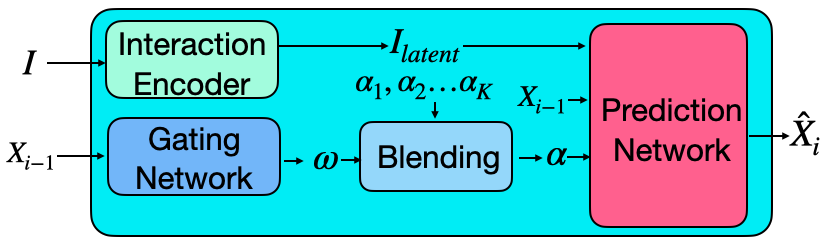}	
	\caption{
	MoE Architecture.
	}
	\label{fig:moe}
\end{figure}

%% file: FIG_shced_sampling.tex
\begin{figure}
	\centering	
	\includegraphics[width=\linewidth]{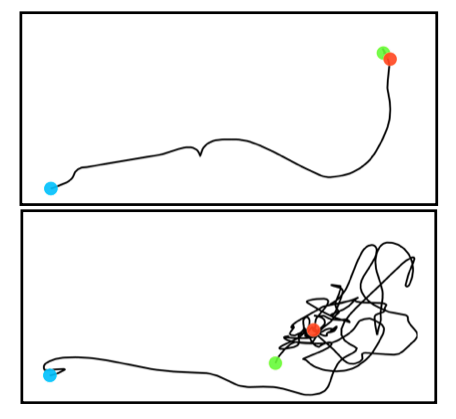}	
	\caption{
	\modelname With Schedule Sampling (Top) and without (bottom). The black line shows the root projection on the $xz$ plane. The blue and green circles denote the root at the first and last frame respectively. The red circle denotes the goal position. Note how \modelname fails to reach the goal without the use of Schedule Sampling.
	}
	\label{fig:sched_sampling}
\end{figure}

%% file: FIG_new_geometry.tex
\begin{figure}[h]
	\centering	
	\includegraphics[width=0.8\linewidth]{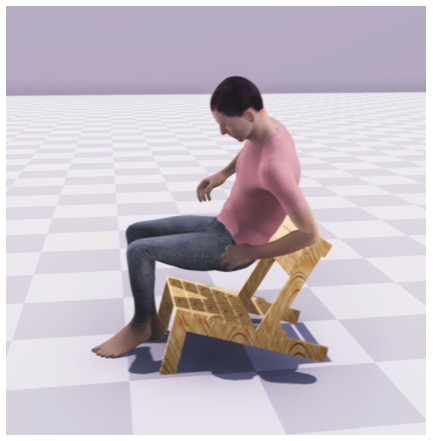}	
	\caption{
	\modelname with significantly different geometry.
	}
	\label{fig:new_goemetry}
\end{figure}